\documentclass[authoryear,preprint,review,3p,times,12pt]{elsarticle}

\usepackage{amssymb}
\usepackage{amsmath}

\usepackage{hyperref}

\journal{In submission to the Journal of Memory and Language}

\begin{document}

\begin{frontmatter}



\title{Less is more: Probabilistic reduction is best explained by small-scale predictability measures} 

\author[UBL]{Cassandra L. Jacobs}
\author[UBP]{Andrés Buxó-Lugo}
\author[EURECOM,PARIS]{Anna K. Taylor}
\author[UBC]{Marie Leopold-Hooke}

\affiliation[UBL]{organization={Department of Linguistics, University at Buffalo},
            city={Buffalo},
            state={NY},
            country={USA}}

\affiliation[UBP]{organization={Department of Psychology, University at Buffalo},
            city={Buffalo},
            state={NY},
            country={USA}}
            
    \affiliation[EURECOM]{organization={EURECOM},
            city={Biot},
            country={France}}

\affiliation[PARIS]{organization={Université Paris - Sorbonne},
            city={Paris},
            country={France}}

\affiliation[UBC]{organization={Department of Computer Science and Engineering, University at Buffalo},
            city={Buffalo},
            state={NY},
            country={USA}}

\begin{abstract}
We explore the cognitive representations and mechanisms that underpin probabilistic reduction in phonetic duration. Recently, massive statistical models of language have been developed that capture long-distance dependencies between words, but the plausibility of these systems as models of human cognition is unclear. Following up on two recent studies, we test in three studies whether the amount of context that these systems use is necessary and/or appropriate when investigating the relationship between language model probabilities and phonetic duration. We establish that the retrieval of multi-word sequences suffice as cognitive units of planning and that retrieval of frequent sequences promotes reduction.
\end{abstract}




\end{frontmatter}

\section{Introduction}


The articulatory dynamics of language production are sensitive to linguistic statistics, such as the frequencies of words, the phrases that those words appear in, and the relative probability of those words given the context \citep{dammalapati2021effects, bell2009predictability,harmon2021theory}.
Familiar words and phrases are assumed to be shorter due to factors like motor practice \citep{bybee2006usage}, or other factors like the general ease of retrieval \citep{arnold2012audience}.
Previous studies have demonstrated that predictability influences articulatory dynamics, such that the more probable a word is, the shorter in duration the words tend to be  \citep{lieberman1963some}.

Words can be predictable in different ways. 
We can both anticipate the future, (e.g., ``carve a...'' is likely to end with \textit{pumpkin}) as well as confirm that a word is supported by later information (e.g., \textit{pumpkin pie}).
Whereas the influence of word frequency on reduction could be explained by motor practice, findings that words are reduced when they are predictable has been argued to support rational production models.
Conversely, lengthening improbable words is believed to afford listeners extra time to interpret the speech signal and minimize the odds of miscommunication \citep{aylett2004smooth,degen2020redundancy,futrell2023information,mahowald2013info,clark_relationship_2025,piantadosi2011word}.
However, probabilistic reduction effects are apparently illusory or contradictory depending on the level of granularity at which probabilities are computed (n-gram versus neural language model), and the type of word (content vs. function; \citealp{bell2009predictability}).
Recent work has argued that large language models are uniquely suited to simultaneously explain probabilistic reduction and repetition reduction \citep{upadhye2025back,clark_relationship_2025}.

These studies have argued that n-grams are fundamentally unable to account for enough context to reliably model reduction relative to large language models (LLMs), which can encode very long sequences of prior context.

However, the cognitive plausibility of large language models as applied toward language production is unclear.
Firstly, it is unknown what knowledge speakers draw on during language production, and how a speaker might efficiently estimate a word's probability in context.
For speakers to know that a particular word will be easily understood because it is partly redundant with the downstream context, and can therefore be reduced  \citep{aylett2004smooth,jaeger2010redundancy}, the language production system must be able to compute such a probability efficiently.
Computing probabilities from a language model typically involves encoding the full, known upstream and/or downstream context to estimate a word's probability.
Alternately, some have applied the fill-in-the-middle procedure of \citet{bavarian2022efficienttraininglanguagemodels} to more efficiently extract words' probabilities, but this procedure does not clearly correspond to any known cognitive processes that speakers engage in to learn language, nor to perform fluent production.
Additionally, as \citet{upadhye2025back} point out, solutions built on beam search or sampling \citep{giulianelli2024generalized} are too expensive to be plausible accounts of production either \citep{meister-etal-2020-beam,pmlr-v97-cohen19a}.

Long-distance context sensitivity of the process of multiword production is at odds with the broad consensus in language production research that speakers do not plan their utterances entirely in advance, and instead plan their utterances relatively incrementally \citep{ferreira2002incremental,futrell2023information,guhe2020incremental,meyer1996lexical,pickering2013forward}.
So, the language production system might plan utterances at a lower level of granularity (e.g., one or two nearby words), or perhaps a larger one when certain kinds of long-distance dependencies are considered \citep{momma2021filling}.
It is widely believed that speakers store multiword sequences in long-term memory and that doing so promotes production fluency \citep{arnon2013more,bannard2008stored,morgan2024productive,siyanova2018production}.
Here we show that local probabilities suffice for the computation of probabilistic reduction, and phrase-based representations are more plausible units of retrieval when considering the computational efficiency of phrase-level retrieval.

To assess these competing accounts, the experiments described in this paper compare and contrast probability estimates of words derived from n-gram and LLM estimates, which are then incorporated as predictors in regression analyses of whole-word phonetic duration. We analyze four spontaneous speech corpora of American English: the Buckeye Corpus \citep{pitt2005buckeye}, the Corpus of Regional African American Language (CORAAL; \citealp{kendall2013speech,kendall2018corpus}), Switchboard-NXT \citep{calhoun2010nxt} of \citet{upadhye2025back}, and the CANDOR corpus \citep{reece2023candor} from \citet{clark_relationship_2025}.
Critically, our analyses include important utterance-level prosodic covariates that are known to affect word durations (\citealp{bishop2018anticipatory,baker2009variability}; Section \S\ref{prosodic}).

We demonstrate the utility of small-context modeling in Section \S\ref{clarkcandor} by showing that LLM probabilities extracted over intonational units, rather than whole conversational turns, better account for the effects of predictability on duration.
We then discuss that language model probabilities may be partially confounded with language model probabilities (Section \S \ref{upadhye-replication}).
We additionally show in Section \S\ref{upadhye-replication} and \S\ref{dialect-corpus-analysis} that statistics from large language models are not necessary to observe probabilistic reduction.
In fact, the results of n-gram analyses are broadly consistent and demonstrate that LLM predictors produce worse model fits to durations than n-gram probabilities.
In Section \ref{effect_size_viz} we then examine the regression models across all datasets and observe that now-common practices for model comparison 
have masked the extremely small and potentially unstable effect size of probabilistic factors on phonetic duration (as discussed in \citealp{bell2009predictability}).
We ultimately provide a verbal account building on multi-word linguistic representations that can explain probabilistic reduction efficiently.

\section{Prosodic Preliminaries}
\label{prosodic}

Words' durations (of course in addition to other dimensions of speech prosody) are sensitive to structural, probabilistic, and utterance-specific factors  \citep{baker2009variability}.
In order to accurately gauge the contribution of probabilistic factors to duration, and determine whether large language models or multiword representations are more appropriate, we discuss these critical covariates.

\subsection{Structural and lexical influences on duration}
``Structural'' factors that are known to affect duration include pauses, syntactic or prosodic phrase boundaries \citep{byrd2003elastic,klatt1975vowel,turk2007multiple,wightman1992segmental}, in addition to lexical information such as number of phonemes and/or syllables \citep{TurkAliceE.1997Tdoa}, a word's phonotactic probability \citep{goldrick2008phonotactic}, and its phonological neighborhood density \citep{gahl2012reduce,gahl2016many}.

\subsection{Utterance length and position confounds on duration}

Word durations have been shown to vary based on utterance planning/production dynamics. 
For example, words produced in isolation in spontaneous speech tend to have the longest durations as speakers may be disfluent and use timing information to hold the floor in a conversation \citep{clark2002using}.
Occasionally, researchers incorporate pause duration or pause location as a control variable \citep{clark_relationship_2025}.
This work instead conceptualizes phrase-final lengthening as a relative position effect, such that words closer to the end of a sentence tend to be longer.

Relatedly, ``anticipatory shortening'' is a pattern where the durations of a syllable in a phrase are negatively correlated with the length of the phrase \citep{bishop2018anticipatory}.
This has been argued to reflect scope of planning effects, as syllable durations throughout the utterance are affected by a factor that is not made explicit until the production is complete. 
Thus, particularly for longer utterances, speech rates may be planned out somewhat in advance based on the approximate number of words or syllables a speaker plans.
Similar findings have been demonstrated in fundamental frequency, such that longer utterances typically start with higher pitch than shorter utterances, suggesting that producers are broadly aware of how long they want to speak \citep{liberman1977stress}. 
In contrast to prior work which has used speaker speech rate \citep{upadhye2025back} or numeric word position as a covariate \citep{clark_relationship_2025}, we use utterance-specific phrase length to account for gross timing patterns.

\begin{figure*}[t]
    \centering
    \includegraphics[width=1\linewidth]{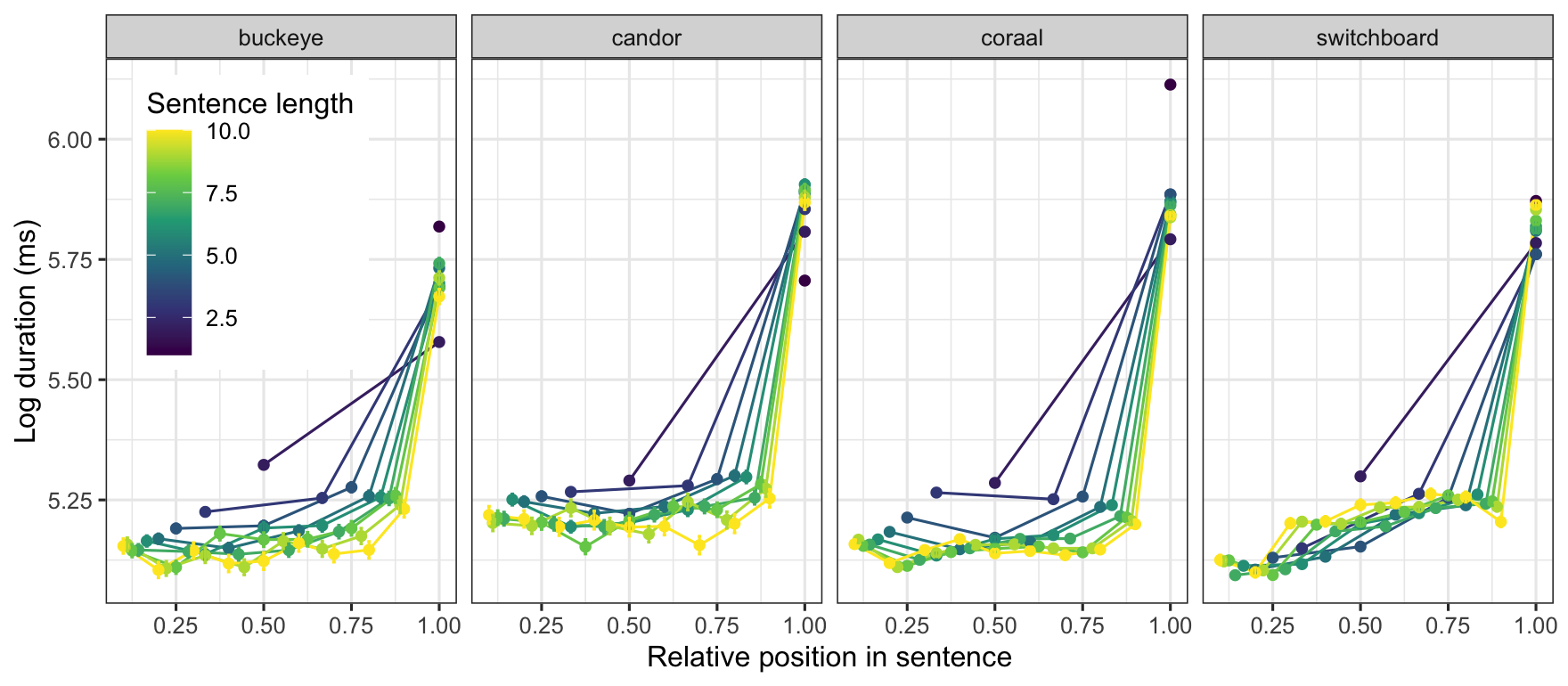}
    \caption{Sentence-final lengthening and anticipatory shortening effects (relative position) on duration across four spontaneous speech corpora. Only utterances comprised of ten words or fewer are visualized; all are analyzed.}
    \label{fig:ASD-effect}
\end{figure*}

To better model utterance length, we chunk utterance boundaries into inter-pause units based on human-annotated pause boundaries.
The resulting prosodic or intonational phrases have been argued to represent a fundamental unit of planning in production \citep{wheeldon2002minimal}.
Segmenting speech along inter-pause intervals and encoding a word's relative position in an utterance produces the canonical anticipatory shortening effects across all datasets except Switchboard, which we show in Figure \ref{fig:ASD-effect}.
All lengths of utterance show the standard phrase-final lengthening effect.

\subsection{Probabilistic factors}
``Probabilistic'' factors that affect reduction include a word's frequency, its forward predictability, and its backward predictability \citep{clark_relationship_2025,bell2009predictability,upadhye2025back}. Words that appear in frequent phrases are also typically reduced \citep{arnon2013more}. In all cases, a word is expected to be reduced in duration when it is more probable in context. Additionally, a word's informativity has a moderating effect, such that words that are typically unpredictable will tend to be less reduced \citep{seyfarth2014word}. 

In the case of two-word sequences, or bigrams, the probability of a word $w$ at position $i$ in context $c$ can be broadly generalized as 

\begin{equation}
c = w_{i-1}, \frac{p(w_{i-1}, w_{i})}{p(w_{i})}
\end{equation} 
for the forward transition probability of encountering the word $w_i$ \textit{next}, and 

\begin{equation}
    c = w_{i+1}, \frac{p(w_i, w_{i+1})}{p(w_{i+1})}
\end{equation} 
for the backward transition probability of having previously encountered $w_i$ given the next word. 

Large language models typically incorporate more context than n-gram probabilities, and can be computed for very large sequences of text, such as an entire conversation \citep{clark_relationship_2025}, but the idea behind the computation is similar.
For language models (including n-grams) with arbitrary memory, instead of conditioning on the immediately preceding word, a word's probability can depend on the complete chain of preceding words in the following way:

\begin{equation}
\label{general_prob_formula}
    p(w_i | c) \propto p(w_i | w_1 ... w_{i-1})
\end{equation}

The probability estimated in Equation \ref{general_prob_formula} is intractable to compute in practice, particularly for longer sequences.
The application of autoregressive language models (e.g., GPT-2; \cite{radford_language_2019}) within psycholinguistics has provided researchers with a means to quantify a word's probability of being the next word given the preceding context through a transformation of the language model's representation of the prior context.
In practice, these probabilities are computed by multiplying the last hidden state with a matrix that produces a vector of activations, which is then scaled to produce a quasi-probability distribution with a softmax (historically sigmoid) transformation \citep{rumelhart1986general}.

\begin{table*}
    \centering
    \begin{tabular}{lcc}
        \textbf{Predictor of interest} & \textbf{Transform} & \textbf{Expected coefficient sign} \\
        \hline
        Utterance length in number of words & log & - \\
        Relative position of word within utterance & inverse log & + \\
        Forward predictability & log & - \\
        Backward predictability & log & - \\
        \hline
    \end{tabular}
    \caption{Expected relationships of critical predictors on whole-word duration.}
    \label{tab:predictions}
\end{table*}

\subsection{Covariate selection}

To control for non-probabilistic factors that could influence phonetic duration, we extracted several predictor variables from previously established norming databases.
For consistency with previous work, we use frequency information from the SUBTLEX-US corpus \citep{brysbaert2009moving}.
We obtain norms for a word's length in number of phonemes from \cite{chee2020consistency}. 
Number of syllables was not found to be a significant predictor in any analysis, and was recommended for exclusion in all model comparisons; therefore, we do not report any models with it included.
Such a finding may be due to the existence of other covariates in the model, particularly by-word random intercepts and other predictors in the fixed effects that encode length, as syllable structure is known to influence word duration \citep{TurkAliceE.1997Tdoa}.
Each word's forward and backward transition probability is computed per dataset by the relevant n-gram or neural language model, described in each section.
Table \ref{tab:predictions} shows the predicted outcomes from these analyses.

To account for phrase-final lengthening, we compute a transform on relative position, itself defined as the position $i$ for a word $w_i$ out of all of the $n$ words in the sentence. $i$ is strictly positive and non-zero. 
We compute an inverse position value that produces higher values the closer the sentence is to the end.

\begin{equation}
    \text{Inverse relative position} = -\log(1 - \frac{i}{n})
\end{equation}

At the end of the sentence ($i=n$), this value is undefined; therefore, we impute the maximum value from other non-initial positions in the other sentences in each corpus to determine this inverse position value.

Where possible, we include random intercepts for each word in a particular part of speech (POS) category (e.g., \textit{extract} as a noun or verb) to account for variation that might arise for homographic words or words used in different sentences 
as well as random intercepts by talker.
For analysis of Buckeye and CORAAL in Section \S \ref{dialect-corpus-analysis}, course-grained POS tags were obtained through spaCy \citep{Honnibal_spaCy_Industrial-strength_Natural_2020}.
We relied on the Penn Treebank POS tags in Section \S \ref{upadhye-replication}, and no POS labels were available for Section \S \ref{clarkcandor}.

\subsection{Data availability}

All data and analyses are available on the Open Science Foundation \href{https://osf.io/bp6ce/overview?view_only=79d0aa0d963f4e3c8cd8bdca52a50953}{(external link)}.

\section{Study 1: Establishing short-range probabilistic reduction in \citet{clark_relationship_2025}}
\label{clarkcandor}

If speakers modulate word durations to make unpredictable words longer and predictable words shorter, this would suggest that speakers are taking into account the needs of the listener.
Some long-distance dependencies, such as a word's discourse status (e.g., given versus new; \cite{fowler1987talkers}), influence word durations as well.
In their analysis of the CANDOR corpus, \citet{clark_relationship_2025} computed a transformation of next-word probabilities known as surprisal and found that words that were less predictable from the upstream context were produced with longer durations. 
They reasoned that the long-distance abilities of the large language model allowed them to account not just for the predictability effects of words at short distances, but also at long distances.

As \citet{clark_relationship_2025} point out, the general length of a turn in conversation is rarely very long, typically under 50 words.
Thus, their use of GPT-2's full context window (up to 1024 tokens) to encode the entire turn is likely overly powerful for estimating a word's probability.
We expect that smaller amounts of context will produce better estimates of a word's probability in context, and therefore better account for variability in word durations.
Thus, in this section, we re-analyzed the CANDOR corpus, but extracted language model probabilities using the probability estimation method of \citet{pimentel_how_2024} only over intonational phrases \citep{wheeldon2002minimal}.
To do so, we re-tokenized CANDOR utterances along known pause boundaries to produce intonational phrases, which shortens the typical lengths of utterances substantially.
As the model estimates were extracted over the same sentences, the two probability measures are highly correlated (Spearman $\rho$ = 0.69).

\begin{figure}
    \centering
    \includegraphics[width=.5\linewidth]{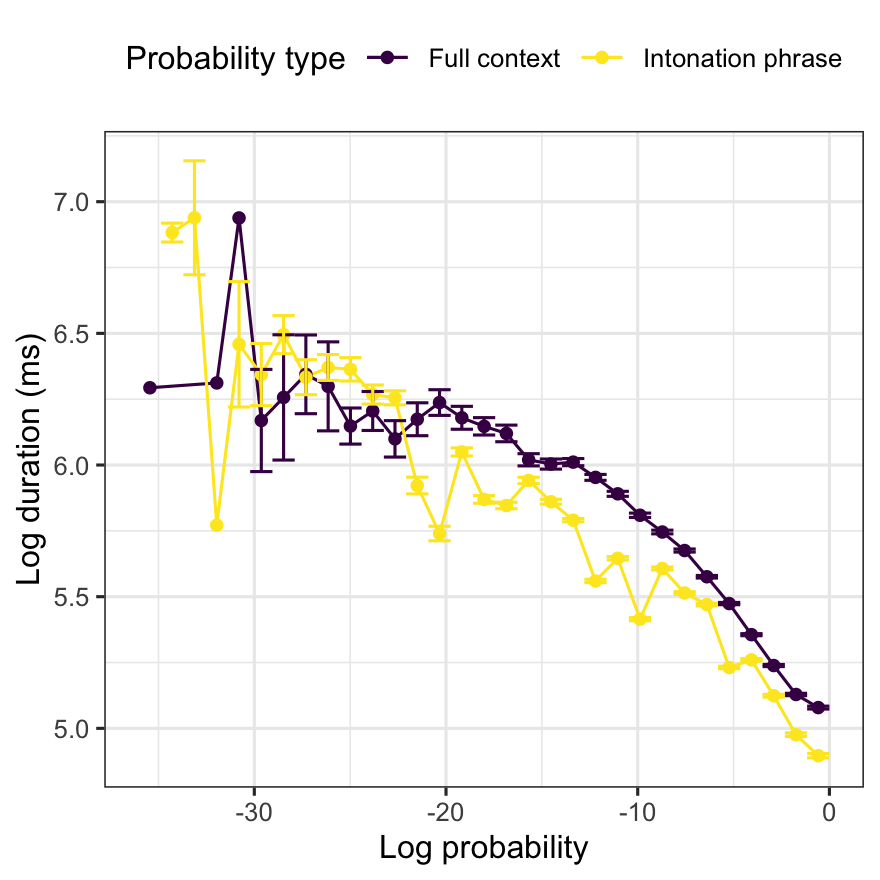}
    \caption{Correlation between log language model probability and words' durations (log ms) for short- and long-range language model probabilities. X axis is binned for visualization.}
    \label{fig:stronger-shorter}
\end{figure}

We find that intonational phrase-level probabilities are more robust predictors of probabilistic reduction.
Critically, the effect of predictability on duration is fully subsumed by the small-context estimate -- there is no contribution to model fit of long-distance probabilities. 
The results of the regression model are included in Table \ref{tab:candor}. 
We visualize the stronger relationship to phonetic duration in Figure \ref{fig:stronger-shorter}.

\begin{table*}
    \centering
    \begin{tabular}{rcrrrr}
\hline
          &            Transform &   Estimate & Std. Error & t value  & p   \\
          \hline
(Intercept)       &   &   0.254 & 0.034  & 7.455 & <.001 \\
SUBTLEX word frequency    & log        &    -0.167  & 0.005 & -28.980 & <.001\\
Number of phonemes      &        &    0.142 & 0.003 & 55.654 & <.001\\
Utterance length in words ($n$)  & log    &  -0.055 & 0.003 & -20.557 & <.001\\
\textbf{Short-range probability} & log & \textbf{-0.021} & \textbf{0.001} & \textbf{-38.126} & \textbf{<.001}\\
Relative position    &  inverse log  & 0.188 & 0.001 & 128.406 & <.001\\
\hline
    \end{tabular}
    \caption{Results of \citet{clark_relationship_2025} re-analysis.}
    \label{tab:candor}
\end{table*}

\section{Study 2: N-gram versus language model predictability and probabilistic reduction in \citet{upadhye2025back}}
\label{upadhye-replication}

Probabilistic reduction may not apply as consistently across all words in the lexicon.
Specifically, function words typically play a more structural or morphosyntactic role in English, and content words are more interchangeable. 
Consequently, some have claimed that function words are retrieved and planned at an earlier stage than content words \citep{garrett1975analysis}.
Consistent with this, \citet{bell2009predictability} describe a dissociation in their analysis of Switchboard, in which content words were only reduced when backward transition probabilities were high, but function words were only reduced when forward transition probabilities were high.

An alternative position to the separate planning of content and function words is that the content-function distinction is somewhat fraught or arbitrary, or mostly reflects lexical frequency \citep{staub2024function}.
Similarly, if speakers retrieve phrase-shaped representations for production, then it is also plausible that probabilistic reduction should apply to both types of words equally.
Such a finding would be consistent with \citet{upadhye2025back}, whose results contradicted \citet{bell2009predictability}, such that forward and backward predictability led to reduction on both content and function words in the Switchboard-NXT corpus.\footnote{\label{footnote-bell} We wish to point out that the content-function distinction in the corpus is somewhat unclear. \citet{bell2009predictability} state, \begin{quote}``The function-word category included discourse markers, conjunctions, existential there, pronouns and other proforms, prepositions, articles, quantifiers, demonstratives, verb auxiliaries, and verb particles. The remainder, nouns, verbs, adverbs, and adjectives, made up the content-word category.''\end{quote} However, an examination of the dataset reveals many words that belong to both categories, a large number of content words are found in the function word category, and pronouns are notably included in the content word category. Future iterations of work in this vein should clarify or correct the content/function distinction.} 
Therefore, in this experiment we replicate their analyses of reduction on content and function word durations.

The Switchboard-NXT data from \citet{upadhye2025back} contains 650,570 observations.
For language model probabilities, we use the provided autoregressive predictors from their custom GPT-2 model \citep{radford_language_2019} trained in both forward and backward directions on the CANDOR corpus\footnote{\citet{upadhye2025back} present many other technological solutions that we find interesting, but these are not the focus of the present work.}.
For analyses of n-gram fluency, we rely on their smoothed bigram estimates derived from the CANDOR corpus, which serve a similar role to the bigram predictors used in \citet{bell2009predictability} and \citet{seyfarth2014word}.
We broadly replicate all of the results of \cite{upadhye2025back} with bigram statistics, such that we observe probabilistic reduction on both content and function words in both forward and backward directions (Table \ref{tab:swbd}).
Importantly, we also find that language model probabilities provide a worse fit to the duration data than bigram probabilities ($\Delta LL = 2562$, $\chi ^2=5123.6$, $p$ < .001).

One concern about these results (raised in Footnote \ref{footnote-bell}) is that the presence of reduction for both Content and Function words could be partly due to errors in the coding of Content or Function word category labels, or due to the somewhat arbitrary distinction between the two. 
We thus conducted an additional analysis without the word class distinction and interaction terms and obtained roughly similar results, again finding that probabilistic reduction in both forward and backward directions is observable with simple bigram statistics.
However, future work should determine the extent to which label noise negatively impacts statistical power to detect differences in patterns of probabilistic reduction across content and function words.

\begin{table*}
    \centering
    \begin{tabular}{rcrrrr}
\hline
          &            Transform &   Estimate & Std. Error & $t$ value  & $p$   \\
          \hline
(Intercept)       &   &   0.138 & 0.022  & 6.096 & <.001 \\
SUBTLEX word frequency    & log        &    -0.124  & 0.004 & -32.227 & <.001\\
Number of phonemes      &        &    0.133 & 0.002 & 86.337 & <.001\\
Utterance length in words ($n$)  & log    &  0.008 & 0.001 & 7.047 & <.001\\
Function words & log & -0.211 & 0.012 & -17.076 & < .001 \\ 
\hline
Forward & & & & & \\
\textit{Content} & log & \textbf{-0.007} & \textbf{0.001} & \textbf{-38.126} & \textbf{<.001}\\
\textit{Function} & log & \textbf{-0.004} & \textbf{0.001} & \textbf{-4.835} & \textbf{<.001}\\
\hline
Backward & & & & & \\
\textit{Content} & log & \textbf{-0.034} & \textbf{0.001} & \textbf{-52.036} & \textbf{<.001}\\
\textit{Function} & log & \textbf{-0.007} & \textbf{0.001} & \textbf{-7.833} & \textbf{<.001}\\
\hline
Relative position    &  inverse log  & 0.092 & 0.001 & 144.307 & <.001\\
\hline
    \end{tabular}
    \caption{Results of \citet{upadhye2025back} re-analysis, excluding \sc other.}
    \label{tab:swbd}
\end{table*}

\begin{figure*}[h!]
    \centering
    \includegraphics[width=1\linewidth]{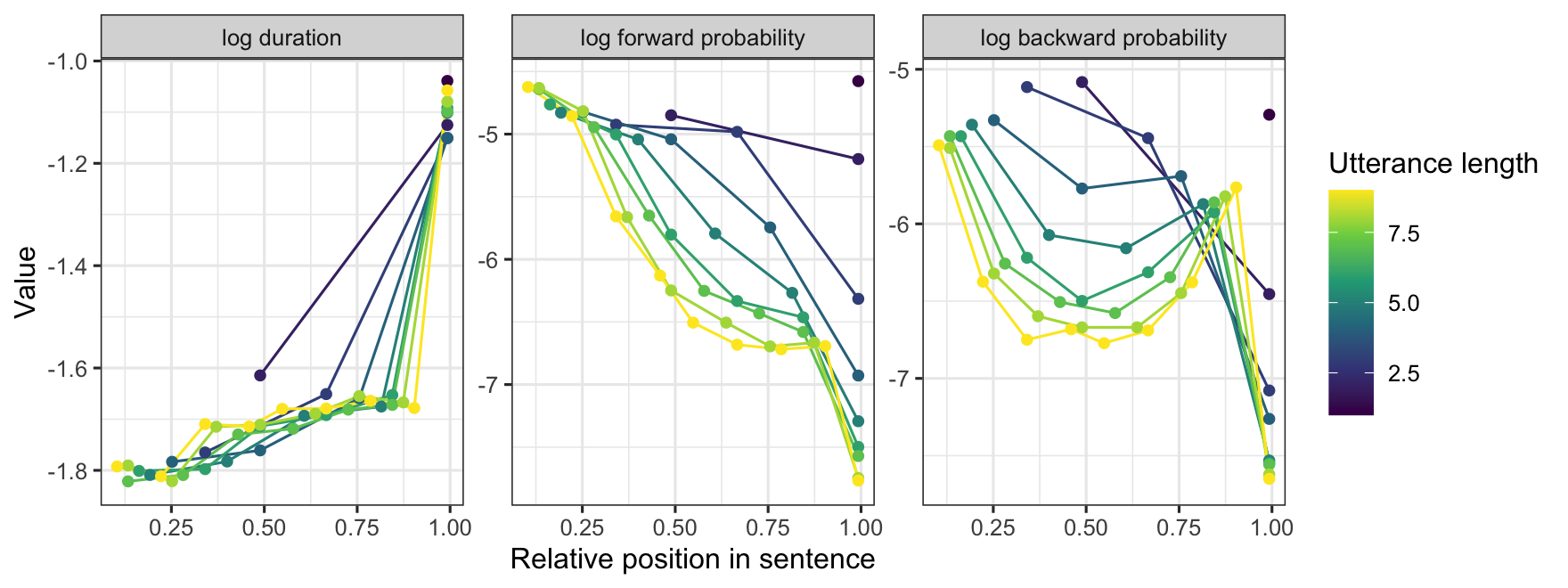}
    \caption{Relationship between utterance length and relative position on language model probabilities and duration in \cite{upadhye2025back}}
    \label{fig:llm}
\end{figure*}

One reason for the poor explanatory power of LLM probabilities might lie in an apparent confound between language model estimates and position within an utterance.
We visualize several complex relationships between position, length, and duration in the Switchboard data in Figure \ref{fig:llm}.
For language model probabilities, forward transition probabilities decrease substantially over positions, with the specific trajectory affected slightly by utterance length.
Backward predictability furthermore shows a complex, highly non-linear pattern in which early, non-initial words and utterance-final words are assigned greater probabilities than utterance-medial words.
Because the durations of words toward the ends of the sentence also increase (see Figure \ref{fig:ASD-effect}), the predictive power of language model probabilities relative to n-grams in \cite{upadhye2025back} may be an artifact of a word's position within an utterance.
Whereas autoregressive language models may encode word position in a sentence more explicitly \citep{haviv-etal-2022-transformer}, such confounds may be less present with the n-gram estimates as they do not explicitly encode position information, a possibility we explore in Section \S \ref{dialect-corpus-analysis}.

In sum, this analysis shows that not only do large language models lack  cognitive plausibility, relative to phrase-based representations that support successful language production \citep{arnon2013more,siyanova2018production}, they also provide worse fits to the data, possibly due to major confounding factors.

\section*{Interim Discussion}

Our re-analysis of \citet{clark_relationship_2025} and \citet{upadhye2025back} suggests that positional information very strongly shapes prosodic structure.
More importantly, it is clear that prosodic structure is partly confounded with probabilities from large language models, and that longer-range estimates of a word's predictability do not provide additional explanatory power for word durations.
Surprisingly, probabilistic reduction is evident in both datasets, suggesting that a word's predictability affects its phonological encoding \citep{aylett2004smooth}.
One major open question is about the generality of probabilistic reduction, both in terms of speech communities and the domain of spontaneous speech under analysis.
The final study explores these questions.

\section{Study 3: Examining probabilistic reduction in smaller and diverse dialect corpora}
\label{dialect-corpus-analysis}

The Switchboard and CANDOR corpus are both corpora containing dialogues between strangers who may be likely to vary word durations to accommodate their addressees.
Spontaneous speech exists in a variety of domains, however, and speakers' own pronunciations might change because they are addressing strangers.
Interview settings have demonstrated probabilistic reduction: \cite{seyfarth2014word} showed that words that are more predictable in the Buckeye corpus are produced with shorter durations based on bigram statistics.
We specifically test whether the same forward and backward predictability-driven reduction relationships are still observed in more casual, self-driven interviews without the pressure of conversational demands.
This final analysis replicates \cite{seyfarth2014word} and extends his analysis to a larger, more diverse corpus that enables us to ask how universal probabilistic reduction is in spontaneous monologue.

Previous work on probabilistic reduction has focused on White Mainstream American English (WMAE), such as that spoken in the Buckeye Corpus \citep{pitt2005buckeye} 
and Switchboard  \citep{godfrey1992switchboard,calhoun2010nxt}. 
This study examines both WMAE and several varieties of African American Language (AAL), as AAL varieties are under-represented in computational psycholinguistics (as elsewhere in natural language processing; \citealp{10887767}).  
Dialects represent a critical test case for modeling probabilistic language processing, because speech communities, and thus the linguistic statistics of those communities, differ along many dimensions.

A major challenge for assessing probabilistic reduction in dialect corpora, and smaller datasets more broadly, is that large language models may not produce high-quality estimates of contextual probabilities, either when using off-the-shelf models, when trained on the corpora directly, or when fine-tuned.
Moreover, while there is no reason to expect that predictability would not shape production in a similar way across dialects, if dialects are modeled by variables extracted from mainstream varieties (e.g., the models used in \cite{clark_relationship_2025} and \cite{upadhye2025back}), then these biases are expected to distort any effects of linguistic statistical knowledge on production.
This experiment therefore explores the feasibility of applying n-gram statistics to probabilistic reduction across smaller and more diverse dialect corpora.

It is important to level the playing field so that LLM probabilities can capture dialect-specific variability in a low-data training regime, particularly since n-gram models estimated over a specific corpus necessarily capture variability intrinsic to a dialect.
\citet{vskrjanec2026language} showed that fine-tuning language models provides one means to capture the knowledge of a person's specific linguistic experience to explain what they will find difficult to process (or produce) as an alternative to training a small-scale neural language model from scratch.
Thus, our experiments in this section compare fine-tuned language models against n-gram models of words' forward and backward predictability.

\subsection{Dialect corpus preprocessing}
In our experiments, we compare and contrast two datasets of spontaneous speech corpora representing Mainstream American English and African American Language. 
For MAE, we use the Buckeye Corpus used in \citet{seyfarth2014word}, which contains 282,565 tokens. 
We compare probabilistic reduction in Buckeye with CORAAL, which is a broad-coverage corpus of African-American Language \citep{kendall2018corpus}. 
Speakers from this corpus come from Atlanta, Georgia; Washington DC at two different time points; Detroit, Michigan; New York City's Lower East Side, Princeville, North Carolina; Rochester, New York; and Valdosta, Georgia and thus represent a range of histories of migration and language contact.
The full CORAAL corpus is much larger than Buckeye, with 1.37 million tokens, and as of this writing is composed of eight distinct dialect groups from across the United States. 

Both corpora clearly demarcate phrase boundaries in the transcripts.
As in Section \S\ref{clarkcandor}, we tokenized the spoken corpora into utterances along pause boundaries, as these roughly correspond to intonational units \citep{bishop2018anticipatory}. 
Buckeye and CORAAL defined pauses somewhat differently, with CORAAL using phonetic criteria (60-70 ms of silence; \citet{kendallthesis09,kendall2013speech}) and Buckeye relying on the intuition of the annotators \citep{pitt2005buckeye,raymond2003analysis}. 
We suspect that these differences are not major, however, as we showed in Figure \ref{fig:ASD-effect} that both corpora demonstrate the expected anticipatory shortening and phrase-final lengthening under our segmentation scheme.

We took gold inter-pause intervals and then produced timestamps for word boundaries for all of the recordings in both corpora.
Since only some dialects had time-aligned transcripts for word boundaries, for consistency we applied a CORAAL-specific language model in the Montreal Forced Aligner (MFA) to obtain durations of words for all recordings \citep{mcauliffe2017montreal}\footnote{The MFA language model was trained on only four dialects: DCA, DCB, PRV, and ROC.}.

\subsubsection{Language model fine-tuning}
As language model probabilities extracted over short sequences appear to be better predictors of word durations (see Section \S\ref{clarkcandor}), 
we chose to fine-tune the Pythia-160M language model \citep{biderman2023emergent} over inter-pause units for both Buckeye and CORAAL experiments using the \texttt{zeldarose} python package \citep{grobol2023zelda}.
Pythia is an autoregressive language model with a high degree of psychometric predictive power, and it has been successfully applied to other psychophysical data such as human reading times \citep{oh2023transformer,oh2024frequency}.
We verified convergence by visually inspecting validation set loss using Tensorboard \citep{tensorflow2015-whitepaper}.
We share fine-tuning parameters in the Appendix. 

In our fine-tuning procedure, we took into account that AAL has many grammatical and lexical features that are relatively unfamiliar to Pythia, relative to WMAE \citep{dayton1996grammatical,mufwene2021african}.
So, for fine-tuning a language model on the CORAAL data only, we merged each utterance with the next to augment the input size and ensure a strong language modeling baseline for both datasets.

\subsubsection{Reverse language model fine-tuning}

To obtain an estimate of a word's backward transition probability, we again start with Pythia-160M as a baseline model, but we reversed the order of all words in each utterance and separately fine-tuned this model on reversed MAE and AAL corpora.
We used the same fine-tuning parameters as for the forward-trained models. 

\subsection{Results}

\begin{figure*}[t]
    \centering
    \includegraphics[width=1\linewidth]{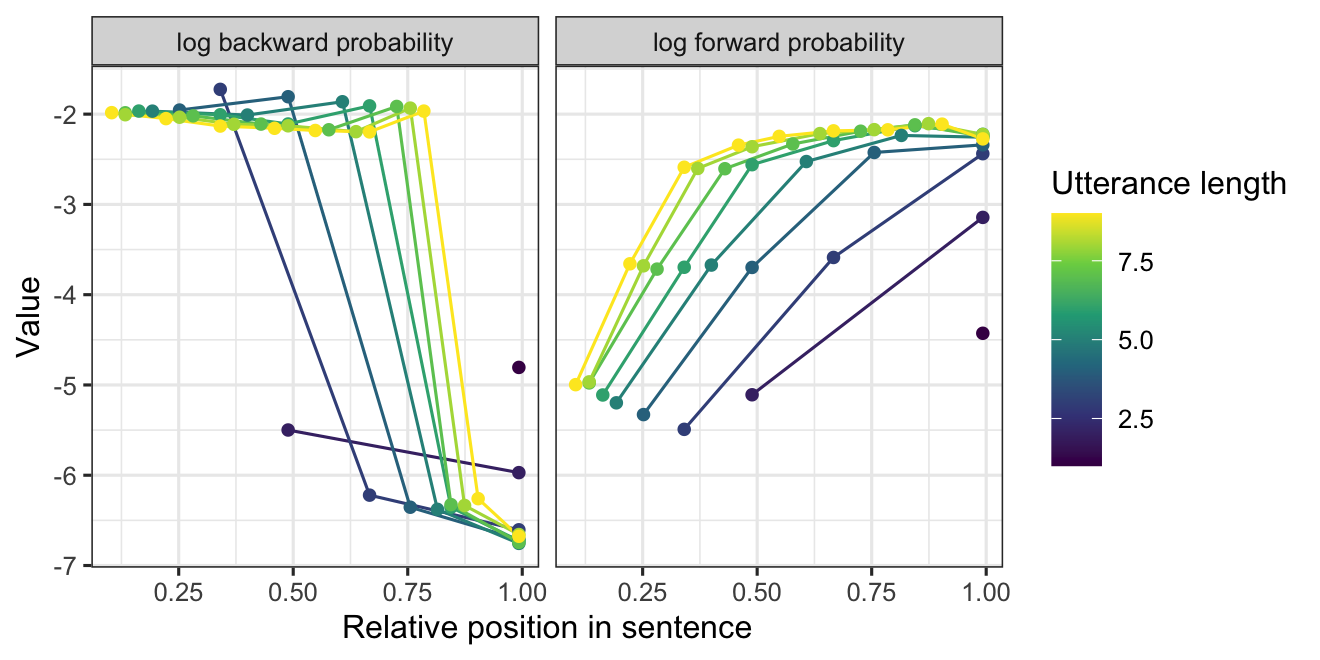}
    \caption{Relationship between utterance length and relative position on trigram conditional probabilities in CORAAL and Buckeye.}
    \label{fig:ngram}
\end{figure*}

The correlation that we observed in the \citet{upadhye2025back} data between neural language model probabilities and relative position persists in these data.
However, for trigram-based estimates of a word's probability in context, we observe a relative lack of correlation, characterized primarily by extreme probabilities at utterance edges, which we illustrate in Figure \ref{fig:ngram}.
Utterance-initial words tend to be relatively improbable in trigram models. 
Forward probabilities typically increase in their predictability after one or two words to a consistent baseline.
Backward probabilities similarly are relatively probable for the early words in an utterance, but these increase to relatively low probabilities by the end of the sentence.

\begin{figure}
    \centering
    \includegraphics[width=1\linewidth]{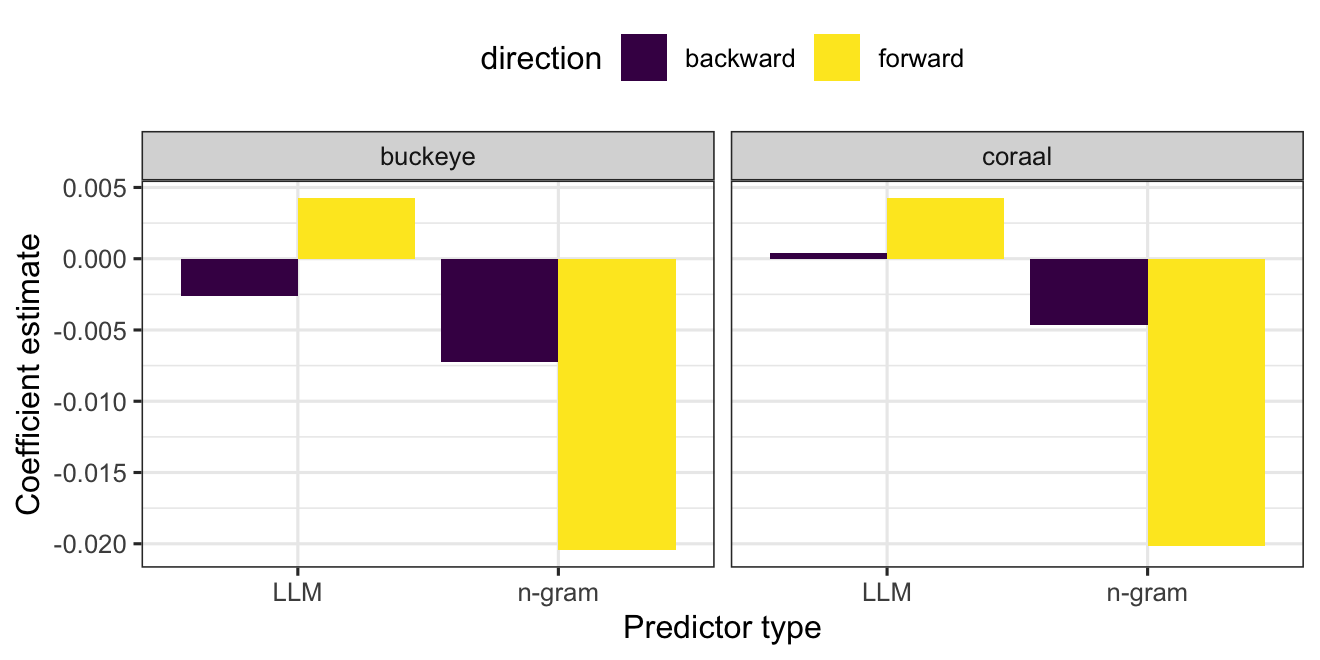}
    \caption{Regression model coefficient estimates for probabilistic reduction by corpus, language model type, and direction for Study 3.}
    \label{fig:effect_probabilities_predictors}
\end{figure}

There is considerable variability across LLM and n-gram predictors in terms of whether they predict probabilistic reduction or not. 
Specifically, in the case of both CORAAL and Buckeye, LLM predictors sometimes predicted lengthening when words are more probable.
On the other hand, all n-gram based predictors show probabilistic reduction.
We visualize the model coefficients from analyses of both corpora and both predictor types in Figure \ref{fig:effect_probabilities_predictors}.
We respectively report the CORAAL and Buckeye regression analyses in Table \ref{tab:coraal} and Table \ref{tab:buckeye} below.

In addition to inconsistent coefficient signs, LLM predictors also proved worse at accounting for variance in duration than trigram estimates ($\Delta LL_{\text{coraal}}$=1515, $\Delta LL_{\text{buckeye}}$=351).
We note that all coefficients are very small, particularly when compared to other predictors in the model, a finding we return to in  the next section.

\begin{table*}
    \centering
    \begin{tabular}{rcrrrr}
\hline
          &            Transform &   Estimate & Std. Error & t value  & p   \\
          \hline
(Intercept)       &   &   0.562 & 0.019  & 29.24 & <.001 \\
SUBTLEX word frequency    & log        &    -0.132  & 0.003 & -38.29 & <.001\\
Number of phonemes      &        &    0.120 & 0.001 & 85.36 & <.001\\
Utterance length in words ($n$)  & log    &  -0.114 & 0.001 & -117.33 & <.001\\
\textbf{Forward transition probability} & log & \textbf{-0.021} & \textbf{0.0002} & \textbf{-67.37} & \textbf{<.001}\\
\textbf{Backward transition probability} & log & \textbf{-0.004} & \textbf{0.0003} & \textbf{-15.72} & \textbf{<.001}\\
Relative position    &  inverse log  & 0.127 & 0.001 & 226.74 & <.001\\
\hline
    \end{tabular}
    \caption{Results of CORAAL analysis with trigram predictability measures.}
    \label{tab:coraal}
\end{table*}

\begin{table*}
    \centering
    \begin{tabular}{rcrrrr}
\hline
          &            Transform &   Estimate & Std. Error & t value  & p   \\
          \hline
(Intercept)       &   &   0.811 & 0.029  & 28.30 & <.001 \\
SUBTLEX word frequency    & log        &    -0.190  & 0.005 & -35.99 & <.001\\
Number of phonemes      &        &    0.129 & 0.002 & 59.42 & <.001\\
Utterance length in words ($n$)  & log    &  -0.100 & 0.002 & -48.99 & <.001\\
\textbf{Forward transition probability} & log & \textbf{-0.020} & \textbf{0.001} & \textbf{-31.62} & \textbf{<.001}\\
\textbf{Backward transition probability} & log & \textbf{-0.007} & \textbf{0.001} & \textbf{-11.49} & \textbf{<.001}\\
Relative position    &  inverse log  & 0.121 & 0.001 & 95.30 & <.001\\
\hline
    \end{tabular}
    \caption{Results of Buckeye analysis with trigram predictability measures.}
    \label{tab:buckeye}
\end{table*}

\begin{figure*}
    \centering
    \includegraphics[width=.9\linewidth]{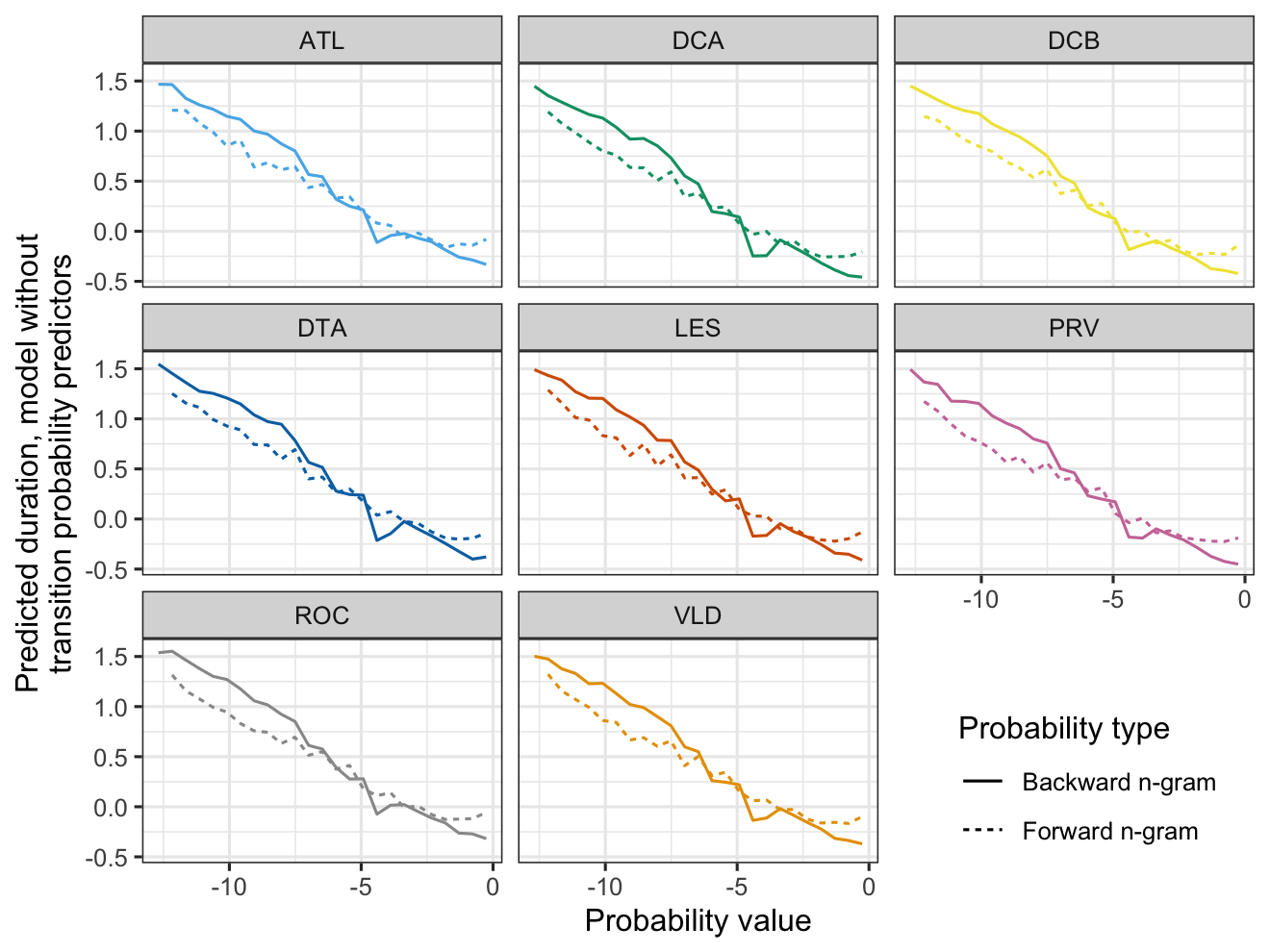}
    \caption{All CORAAL dialects show the same effect of predictability on duration.}
    \label{fig:dialect-differences-coraal}
\end{figure*}

We also attempted to determine whether the different dialect groups in CORAAL varied in the extent to which probabilistic reduction was evident.
A model that included dialect did not converge, and reduction effects were stable across all dialects. We visualize the stability of this pattern in Figure \ref{fig:dialect-differences-coraal}.

\section{Effect size of probabilistic reduction}
\label{effect_size_viz}

As we report, many previous studies have reported contradictory results about the relationship between a word's predictability and its duration, including our own comparison between LLM and n-gram based predictors.
One plausible explanation for inconsistencies across studies of probabilistic reduction effects likely lies in the absolute size of the effect. 
This potential explanation is corroborated by the finding that LLM probabilities fail to consistently produce negative coefficients. 
In this section, we illustrate how the coefficient estimates of probabilistic reduction are dwarfed by the estimates of other factors, even though all probabilistic predictors are highly significant and model comparison strongly supports inclusion of these predictors in the full model.

\begin{figure}
    \centering
    \includegraphics[width=.6\linewidth]{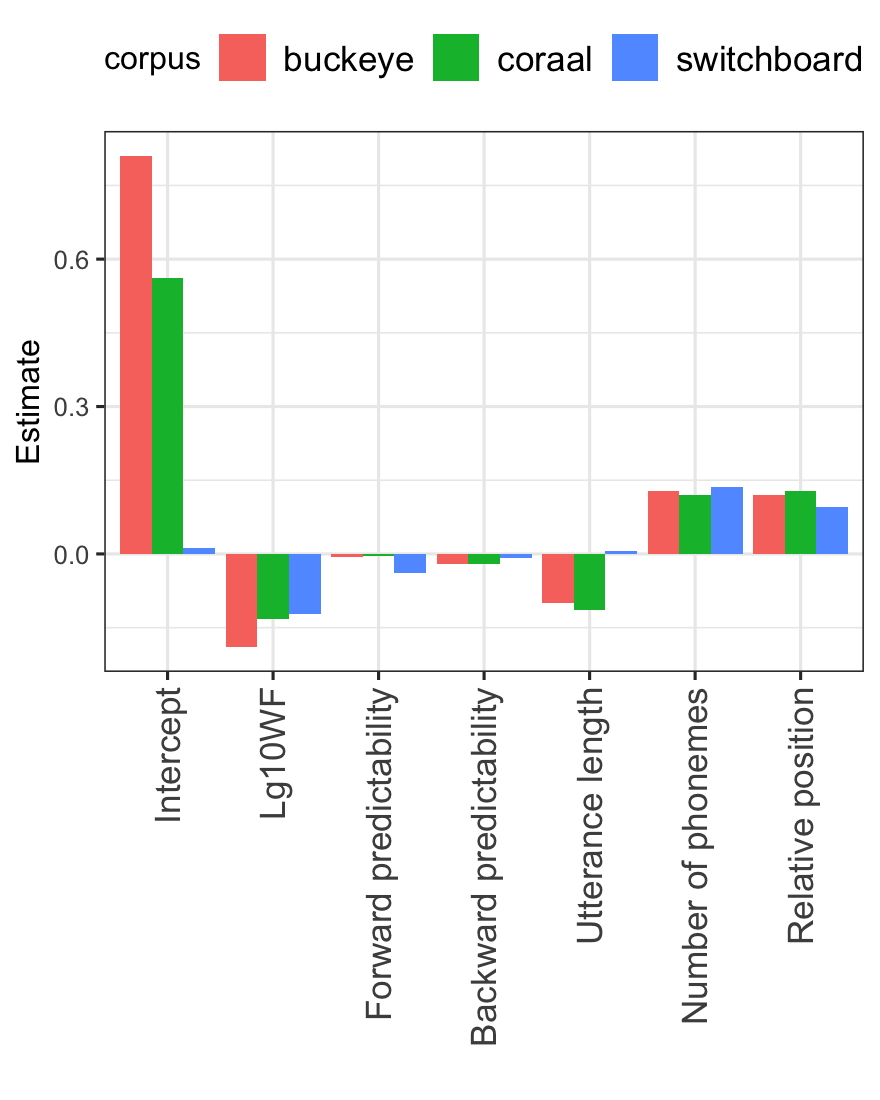}
    \caption{Coefficient estimate magnitudes of control variables and probabilistic predictors of interest.}
    \label{fig:ngram_effect_size_exps}
\end{figure}

When considering coefficient $t$ values as a proxy for standardized effect sizes, many predictors, particularly ``garden variety'' prosodic ones such as the relative position of a word in an utterance, can be an order of magnitude larger than these contextual probability estimates.
While all n-gram analyses are consistent with each other in the sense that probabilistic reduction is observed across all experiments, the effects themselves could be easy to miss, and studies may be underpowered, particularly depending on regression model parameterization.
Figure \ref{fig:ngram_effect_size_exps} illustrates the results of the Switchboard re-analysis from Section \S \ref{upadhye-replication} alongside the results of the CORAAL and Buckeye analysis from Section \S \ref{dialect-corpus-analysis}.

Inconsistency in results is also likely obscured due to changes in statistical reporting practices.
It is now commonplace in computational psycholinguistics to justify the inclusion of probabilistic predictors through model comparison, following early work by \citet{bell2009predictability}.
However, one shortcoming of this approach is that the field's preferred measure, the difference in model log-likelihood ($\Delta LL$), does not include sign information, nor does it transparently convey the absolute value of the coefficient either.
That is, it is possible to report a major improvement in model fit that does not necessarily support claims of probabilistic reduction, and such an effect may be quite small.
Reliance on $\Delta LL$ likely exacerbates the difficulty of finding a significant true effect of probability on phonetic duration, which appears to be highly sensitive to regression model configuration and conditions.

On a more practical note, the effect sizes of probabilistic reduction that we demonstrated in this paper are so small as to be almost imperceptible.
While the significance level or model comparison differences might appear large, the empirical effect sizes of probabilistic reduction suggest speedups of mere single-digit milliseconds between very probable and very improbable words. 
For the listener, such an effect likely does not confer many benefits to language comprehension or spoken word recognition.
For the speaker, shortening words' durations may provide some benefit in terms of resource conservation, but this link is less than clear, and further experimentation is needed to test this alternative explanation. 

\section{General Discussion}
Efficiency is critical for fluent production.
If a speaker is attempting to assess the probability of a word given the context in order to optimize the duration of that word, then they must be able to quickly instantiate or retrieve prior knowledge of linguistic statistics.
In this work, we demonstrated that phrase-based representations provide a plausible representation to produce probabilistic reduction in duration.
Despite claims by \citet{clark_relationship_2025} and \citet{upadhye2025back} that the power of large language models to remember very long contexts uniquely enables them to capture probabilistic reduction better than any other alternative, we find that such claims are emphatically not supported by the data.
Throughout all of our experiments, it is clear that less context produces higher-quality estimates of speech production data, and more consistently.
That is, once prosodic timing variables specific to whole utterances are considered, particularly word position and utterance length, LLM-based predictors provide worse fits to duration data, potentially due to a number of confounds between position, length, and language model probability estimates.

In addition to being poor explainers of phonetic duration, the application of LLMs to the study of probabilistic reduction in general poses several ethical challenges.
First, many language models are typically taken ``off-the-shelf'', without regard to the nature of the web origins of their training data.
In addition to domain differences between spontaneous speech and written language, LLMs that superimpose the statistics of the dominant variety on marginalized varieties will ultimately perpetuate hegemonic data practices \citep{birhane2021impossibility,birhane2022values}. 
Fine-tuning or training from scratch may serve as one mitigation strategy to reduce the biases caused by these decisions, but clearly this is not necessary or sufficient to show probabilistic reduction.

Such findings provide strong evidence, counter to the zeitgeist, that small-scale, relatively incremental production, is critical for capturing phonological encoding effects.
We argue that the consistent explanatory power of phrase statistics, and smaller-scale large language model probability estimates, suggests that if speakers are in fact using some estimate of a word's probability in context during the phonological encoding process, they are doing so in a relatively local way.\footnote{The reader should note that the judgment of next-word probabilities need not rely on exactly the same knowledge or representations as the algorithm that produces an utterance from a particular message (c.f., \citealp{degen2020redundancy}).}
What we observe as effects of rational production, in which speakers purportedly reduce words that are highly predictable, may instead reflect other processes \citep{cohenpriva26}.
For example, we think it is reasonable that speakers retrieve phrases and prosodic knowledge about those phrases from memory somewhat incrementally during production, which could lead to reduction that appears to be sensitive to conditional probabilities, but which instead reflects something like bigram or trigram frequency \citep{bannard2008stored,arnon2013more}.
Indeed, for many practical purposes, these variables might be roughly interchangeable \citep{bell2009predictability}.

In practical terms, we wish to illustrate to the reader exactly how much more efficient language production can be at the level of multiword sequences, relative to knowing the entire content of the message through the end of the utterance.
In order to compute the probability of the final word given the full prior context, the large language models that have been applied in \cite{clark_relationship_2025} and \cite{upadhye2025back} make use of approximately 160 million trainable parameters.
If we consider that the models must also run in the backward direction, then capturing forward and backward probabilities requires 320 million parameters.\footnote{There are some technological solutions to this requiring fewer parameters and computations. However, this approach has questionable cognitive plausibility \citep{bavarian2022efficienttraininglanguagemodels,upadhye2025back} and these models are nevertheless still quite large, at least 100M parameters.}
If we consider then that each utterance must be encoded in full, then the language model must compute the probability of each token in an utterance in order, as it produces a latent representation of the context thus far for every single token in the input.
Typically, at every timestep (every individual token), the model must predict 32k to 50k distinct vocabulary items (or possibly slightly less; \citealp{oh2024leading,pimentel_how_2024}).
Thus, every token possibly transitions into every other token under a neural language modeling scheme.
Concretely, computing the probability of the next word for a sentence as short as 5-6 words will require, at minimum, 800 million computations and a speaker must consider 260,000 possible next tokens.
Note that this problem is not solved if we simply restrict the space to words that are sufficiently active (e.g., \citealp{dell1986spreading}) because these language models obligatorily compute the probabilities of all potential continuations -- restriction of the search space only takes place \textit{after} computing next-token probabilities.

In contrast, n-gram representations are very sparsely connected \citep{evert2005statistics}, and so transition probabilities into and out of a word can be efficiently computed. 
For example, in the CORAAL corpus, the most frequent, best-connected word (``and'') appeared at the end of 14,825 trigrams, and at the beginning of only 13,035 trigrams.
This degree of connectivity suggests that even the worst-case scenario for two- and three-word phrases still represents only a third of the search space as large language model vocabularies require.

\begin{figure}
    \centering
    \includegraphics[width=0.66\linewidth]{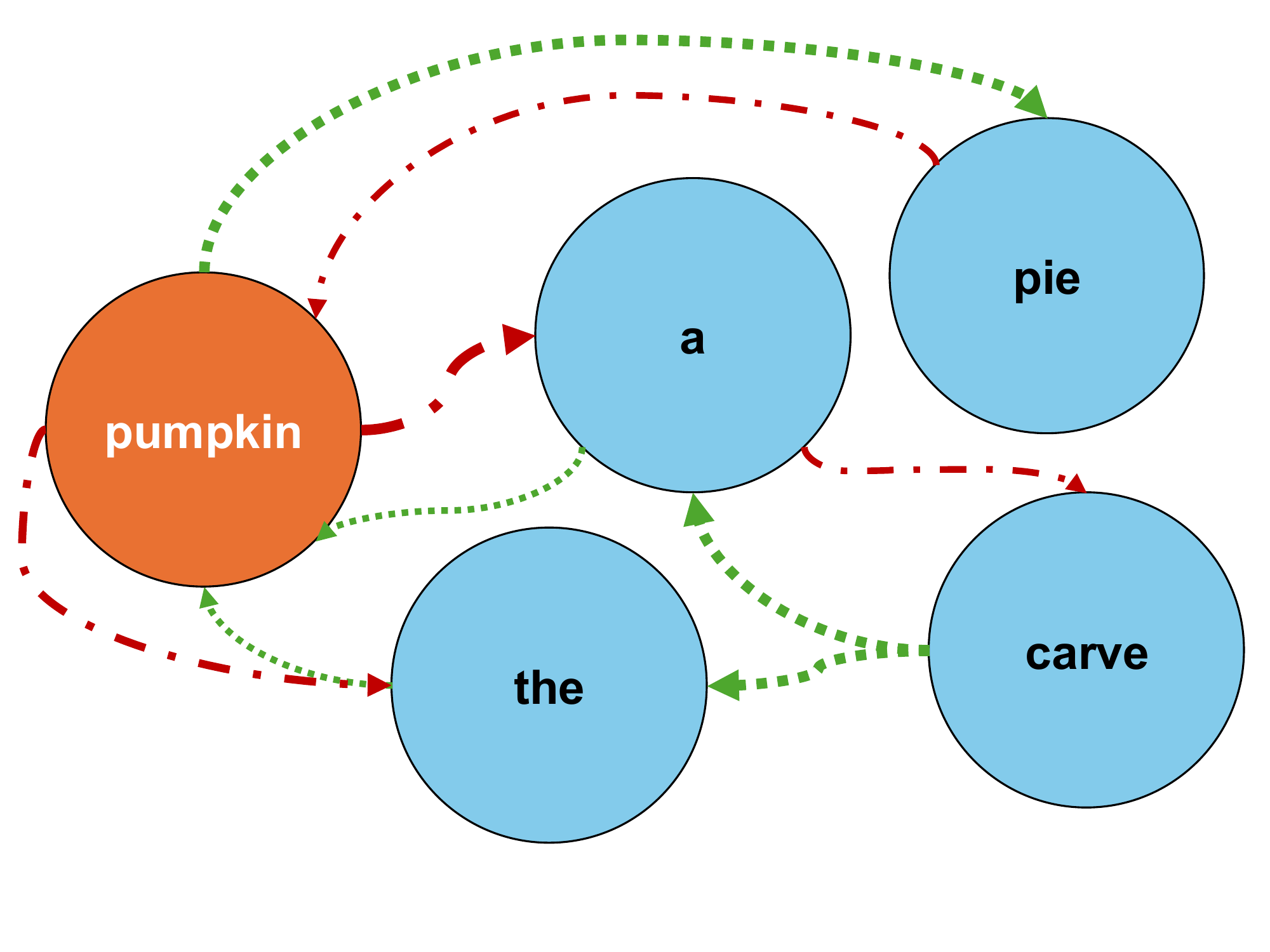}
    \caption{Graphical schematic of phrase associations in long-term memory for the words "carve", "a", "pumpkin", "the", and "pie."}
    \label{fig:transitions}
\end{figure}

\citet{jacobs2017phrase} provided an account of multiword sequence production that was informed by theories of episodic memory that we adapt here to explain predictability effects.
Under their account, one potential representation of phrases in a speaker's mental lexicon could be instantiated as a kind of search through memory; words that are associated with more of the same contexts together will tend to be co-activated and, in the event of retrieval of one word, can cue each other during production.
We can construct memory in a network structure by linking pairs of words (or larger n-grams) to each other.
We illustrate this in Figure \ref{fig:transitions}.
Links $(w_i,w_j)$ are instantiated in both forward (green) and backward (dark red) directions, and the weights associated with these can be extracted or estimated by tabulating an approximate number of shared temporal contexts between two words \citep{howard2002distributed}.
For example, because the word \textit{pumpkin} is likely to follow \textit{carve a}, it has a high backward transition probability weight; \textit{pumpkin} is similarly predictive of \textit{pie}, and so the weight there is also substantial.
Because the network of associations is sparse, estimating transition probabilities from these weights is inexpensive and can be accomplished by instantiating a kind of look-up table.
Obtaining estimates by this route is much more efficient than the LLM approach because it drastically narrows the search space.
Then, the weights can serve as the input to some rational process that might otherwise be approximated using large language models to determine a word's duration \citep{aylett2004smooth,degen2020redundancy,futrell2023information}.

Overall, a phrase-based approach to understanding the production of spontaneous speech provides a clear view of probabilistic reduction.
This paper demonstrated that phrase-level representations and probabilities derived from statistical knowledge of n-grams provides a more plausible and more predictive account of production fluency.
The present work lays the foundation for computational cognitive models of the degree to which we tailor our utterances to suit a particular communicative context, and whether and how we do so in a way that maximizes the odds of successful communication.


 \bibliographystyle{elsarticle-harv} 
 \bibliography{latex/custom}



\newpage

\appendix

\section{Appendix}
\label{sec:appendix}

Fine-tuning parameters used in experiments in Section \S\ref{dialect-corpus-analysis}: 
\begin{table}[h]
    \centering
    \begin{tabular}{cc}
        \hline
        max\_epochs & 40 \\
        batch\_size & 64 \\
        betas & (0.9, 0.98) \\
        epsilon & 1e-8 \\
        learning\_rate  & 1e-4 \\
        lr\_decay\_steps & 1048567\\
        warmup\_steps & 1024 \\
        weight\_decay & 1e-5\\
        \hline
    \end{tabular}
    \caption{Fine-tuning parameters used in all experiments}
    \label{tab:finetuning}
\end{table}

\end{document}